\documentclass{article} 
\usepackage{iclr2026_conference,times}

\usepackage{amsmath,amsfonts,bm}

\def\eqref#1{equation~\ref{#1}}

\def\1{\bm{1}}

\DeclareMathAlphabet{\mathsfit}{\encodingdefault}{\sfdefault}{m}{sl}
\SetMathAlphabet{\mathsfit}{bold}{\encodingdefault}{\sfdefault}{bx}{n}

\usepackage{hyperref}
\usepackage{url}
\usepackage{graphicx}

\title{Causal Evidence of Stack Representations in Modeling Counter Languages Using Transformers}

\author{Nishit Singh\\
Birla Institute of Technology and Science, Pilani\\
\texttt{f20221317@pilani.bits-pilani.ac.in} \\
}

\iclrfinalcopy 
\begin{document}

\maketitle

\begin{abstract}
Formal languages have proven to be effective conduits to understand the inner mechanisms of transformers. Past work has shown that transformers trained on next-token prediction over counter languages learn representations consistent with an underlying stack structure. Beyond representational analysis, this paper investigates the causal role of these representations. Linear probes are trained to predict the stack depth at each token from the model’s hidden states, and a principal representation direction is extracted from the probe. Ablation of this direction from the model causes sequential accuracy to collapse to near 0\%, providing strong empirical evidence that the stack representation is not just learned, but is causally necessary for model performance. 
\end{abstract}

\section{Introduction}

\textbf{Mechanistic interpretability} aims to understand the blackbox-like behavior of language models (LMs) by, among other things, identifying features \citep{elhage2022toymodelssuperposition}. A feature is a human-interpretable property of the model's activation on specific inputs. We use probing classifiers to link these activations to properties. Formally, let \(f : x \rightarrow \hat{y}\) be a language model trained on auto-regressive next-token prediction of some language. A probing classifier \(p : f_l(x) \rightarrow \hat{z}\) returns a property \(z\) from the embeddings of the LM at some layer \(l\). The probing classifier is trained and evaluated on a dataset \(D_P = \{f_l(x), z^{(i)}\}\) which constitutes the embedding and the property of the sequence at that layer \citep{belinkov-2022-probing, hewitt-liang-2019-designing}. While probing classifiers are capable of extracting the latent representations of the language model, \textit{they do not make any statements about causality}. The computational primitives developed during training may or may not be utilized in the computation undertaken by these models.

\textbf{Formal languages} prove to be effective testbeds for performing experiments in interpretability, owing to their mathematically rigorous formulation \citep{bhattamishra-etal-2020-ability, hahn-2020-theoretical}. All algorithmic tasks can be reduced to a language within a specific class of formal languages \citep{DBLP:journals/corr/abs-1901-03429}, making this line of inquiry worthwhile. \cite{tiwari2025emergent} empirically show the emergence of stack representations in language models trained on formal languages like Dyck-1 and Shuffle-k. They explicitly do not make any comments on the causality of these stack representations.

This paper builds on previous research by investigating the causal role of stack representations. A probe is trained to identify stack depth and the corresponding component from the model’s hidden states is ablated during inference. This intervention leads to a near-complete collapse in sequence prediction accuracy, suggesting that the learned stack representation is causally necessary for the model’s internal computation.


\section{Setup}
\label{setup}
\subsection{Counter Language Modeling}
We carry out our experiments over the languages Dyck-1 and Shuffle-k \citep{bhattamishra-etal-2020-ability} for \(k = 2,4,6,8\). Following the formalism by \cite{tiwari2025emergent}, Dyck-1 can be defined as a language over alphabet \(\Sigma = \{ '(', ')'\}\) with production rules: \[
S \rightarrow
\begin{cases}
\epsilon \\
(S) \\
SS
\end{cases}
\]
Shuffle is a binary operation over two strings which interleaves two strings in all possible ways. Formally, for strings $u, v \in \Sigma^*$, the shuffle $u \odot v$ is defined inductively as:
\[
u \odot \epsilon = \epsilon \odot u = \{u\}
\]
\[
au \odot bv = \{a w : w \in u \odot bv\} \;\cup\; \{b w : w \in au \odot v\}
\]
for $a, b \in \Sigma$.
For languages $L_1, L_2 \subseteq \Sigma^*$, define:
\[
L_1 \odot L_2 = \bigcup_{u \in L_1, v \in L_2} u \odot v.
\]
Let $\{\Sigma_i\}_{i=1}^k$ be pairwise disjoint alphabets, and let $L_i \subseteq \Sigma_i^*$ be Dyck-1 languages over each alphabet. Then, the Shuffle-$k$ language is defined as:
\[
\mathrm{Shuffle}_k = L_1 \odot L_2 \odot \cdots \odot L_k.
\]
Intuitively, $\mathrm{Shuffle}_k$ consists of all strings formed by interleaving $k$ independent Dyck-1 strings while preserving the order of symbols within each string.

\subsection{Transformer Architecture}
An encoder-only transformer is trained with a causal attention mask for next-token prediction over Shuffle-k strings. The vocabulary consists of \(k\) bracket pairs, one padding token and one \texttt{BOS} token. Given a prefix, the model predicts the set of valid continuations as a \(k\)-hot vector. The model uses embedding dimension \(d_{embed} = 32\), hidden dimension \(d_{model} = 64\), a single layer with 4 attention heads, feedforward dimension \(d_{ffn} = 64\), and learned positional encodings. The decoder is a linear layer with sigmoid activation, and training uses MSE loss with RMSProp (lr \(5 \times 10^{-3}\), batch size 32, over 10,000 sequences (lengths 2–50, 80/20 training/validation split).

 Positional accuracy \(A_{pos}\) is defined as the fraction of valid token positions where the model's predicted set is equal to the ground truth set. \[\text{A}_{\text{pos}} = \frac{1}{|\mathcal{M}|} \sum_{(b,t) \in \mathcal{M}} \mathbf{1}\left[\hat{y}_{b,t} = y_{b,t}\right]\]
Similarly, sequential accuracy of a string is defined as the fraction of strings which are valid in their entirety. One wrong token deems the whole string invalid, and thus this is naturally a more sensitive metric. 
\[\text{A}_{\text{seq}} = \frac{1}{B} \sum_{b=1}^{B} \mathbf{1}\left[\forall\, t \in \mathcal{M}_b : \hat{y}_{b,t} = y_{b,t}\right]\]
\[\text{where we define } \hat{y}_{b,t} = \mathbf{1}\left[\sigma\!\left(\text{decoder}(h_{b,t})\right) > 0.5\right]\]

This language model achieves perfect validation accuracy in Shuffle-\(k\) for \(k = 2,3,..,8\) by the 25th epoch, confirming that all representations analyzed in this work belong to a fully converged model. 

\subsection{Probe Design}

Shuffle-\(k\) can be represented as \(k\) different stacks, each pertaining to a singular bracket type. An open bracket of type \('i'\) (\(open_i\)) increases the depth of the \(i\)th stack (\(depth_i\))by 1, and a closing bracket \(close_i\) decreases the stack depth by 1. The validity condition of the string is to have \(depth_i=0\)  \(\forall i\). 

Now, for a string \(x = (x_0, x_1,..., x_{n-1})\) the transformer produces, for a layer \('l'\) at token position \('t'\) a hidden state \(h_t^{(l)}\) corresponding to \(c_i(t)\), the depth of the \(i\)th stack at token position \('t'\). We now have:
\[D_P(l) = \{ h_{t}^{(l)}, c_i(t)\}\] 
which is used as the training dataset for the probe. We construct a \textit{linear classifier probe} \citep{belinkov-2022-probing} :

\[f(h) = Wh + b, W \in \mathbb{R}^{C \times d_{model}}, b \in \mathbb{R}^C\] 
where \(C = \) number of depth classes. Trained with cross entropy loss: 
\[\mathcal{L} = -\sum_{(h,c) \in \mathcal{D}_{\text{train}}} \log \frac{e^{f(h)_c}}{\sum_{j=0}^{C-1} e^{f(h)_j}}\] 
using Adam with learning rate \(10^{-3}\), batch size 32, for 10 epochs, with Xavier uniform weight initialization. A \textit{control probe}, as described by \cite{hewitt-liang-2019-designing}, is trained on the dataset \(D_{ctrl}=\{ h_{t}^{(l)}, \pi (c_i(t))\}\) \footnote{where \(\pi(x)\) denotes a random perturbation of the training labels.} and validated on the original validation set. Selectivity is then defined as the difference in the accuracy of the experimental probe and the control probe. High selectivity is an indicator of the stack depth genuinely being encoded in the hidden space vectors. As reported by \cite{tiwari2025emergent}, selectivities for linear probes for Shuffle-k are presented in Table \ref{tab: selectivity}.

\begin{table}[t]
\caption{Some Selectivities of Linear Probes for Shuffle-K}
\begin{center}
\begin{tabular}{ll}
\multicolumn{1}{c}{\bf Value of k}  &\multicolumn{1}{c}{\bf Selectivity}
\\ \hline \\
\(k = 1\), Stack 1         &48.8\%\\
\(k = 2\), Stack 1             &68.7\%\\
\(k = 4\), Stack 1             &83.6\%\\
\(k = 6\), Stack 2           &85.8\%\\
\label{tab: selectivity}
\end{tabular}
\end{center}
\end{table}

\subsection{Intervention}
The intervention follows the activation patching paradigm of \cite{nanda2023progressmeasuresgrokkingmechanistic} and \cite{meng2023locatingeditingfactualassociations}, where targeted manipulation of specific representation directions provides evidence of causal necessity. A linear classifier probe lets us extract the trained probe weight matrix \(W \in \mathbb{R}^{C \times d_{model}}\). Performing singular value decomposition \[W = U\Sigma V^{\dagger}\] 
\[\text{where } U \in \mathbb{R}^{C \times C}\text{, } \Sigma \in \mathbb{R}^{C \times d_{model}} \text{, and }V \in \mathbb{R}^{d_{model} \times d_{model}}\] 
Defining \(\textbf{w}= V_{[0]} \in R^{d_{model}}, ||\textbf{w}|| = 1\) as the top right singular vector of \(W\), we move forward with ablating the direction of \(\textbf{w}\). For each hidden state \(h_t^{(l)}\) at token \(t\) in layer \(l\), living in \(\mathbb{R}^{d_{model}}\), we perform the following operation: \[h' = h-\alpha(\textbf{w}\cdot h)\textbf{w}, \alpha \in [0,1]\]
We also ablate random directions \(\textbf{w}\sim Uniform(\mathcal{S}^{d_{model}-1})\). The ablated states are only passed through the decoder head, the transformer's internal computations are not re-run.

\section{Results}
\label{headings}

\begin{figure}[h]
\begin{center}
\includegraphics[width=0.8\linewidth]{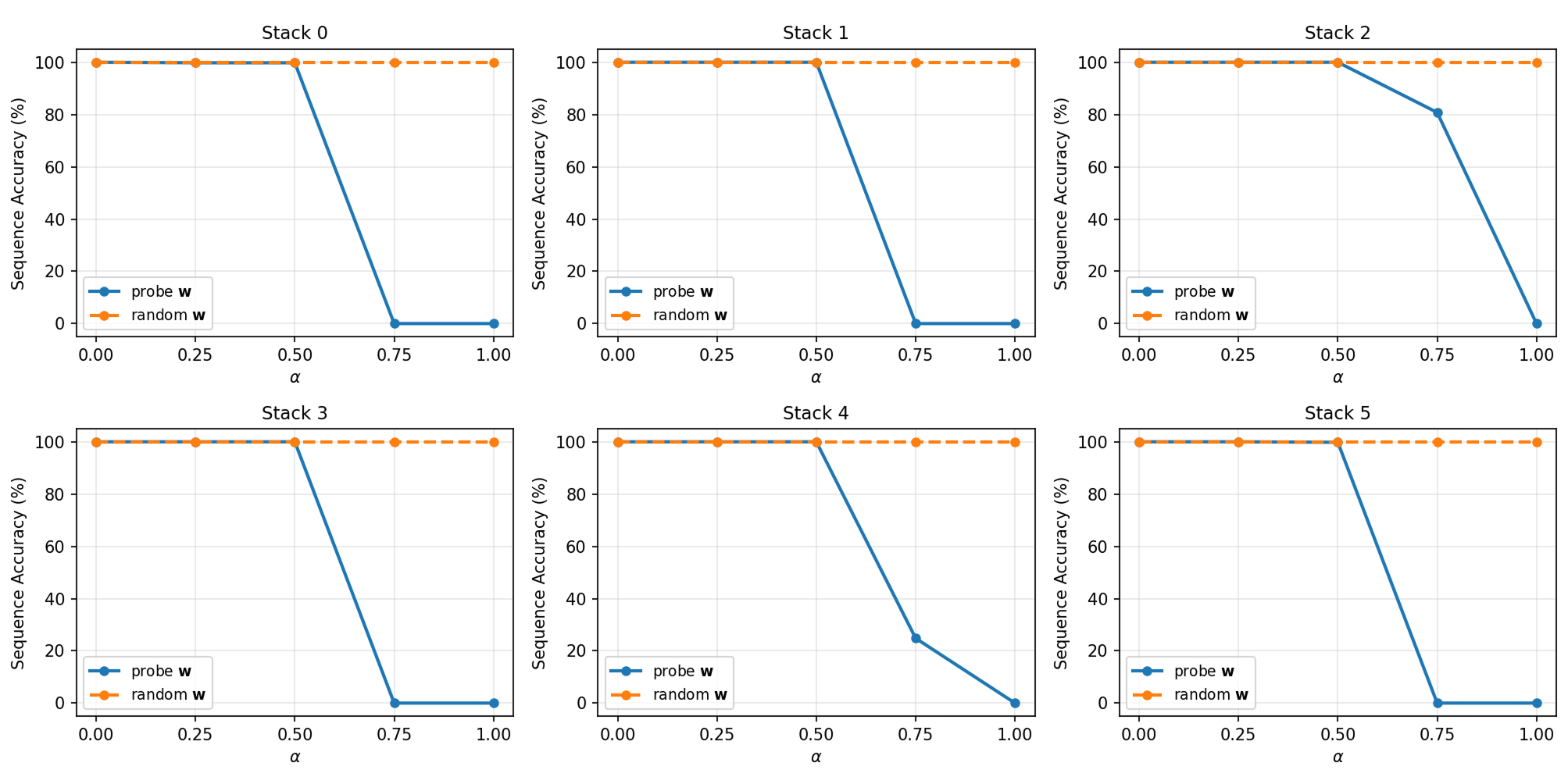}
\end{center}
\caption{Sequence accuracy as a function of ablation strength $\alpha$ for Shuffle-6. Accuracy collapses from targeted ablation, random ablation keeps accuracy intact.}
\label{fig:seq_acc}
\end{figure}

The experiment was run\footnote{Results for \(k=2,4,6,8\) are provided in the Appendix A.} on Shuffle-\(k\) for \(k = 6\). Figures \ref{fig:pos_acc} and \ref{fig:seq_acc} show the ablation of \(\textbf{w}\) from each stack and sweeping \(\alpha = \{0, 0.25, 0.50, 0.75,1\}\) causes the positional accuracy to dip, and sequential accuracy to collapse. The heightened sensitivity of sequential accuracy is expected, since modest per-position accuracy drops compound to severe drops in sequential accuracy. Furthermore, ablating random directions has no effect on the accuracies of the model. Building upon the discovery of emergent stack representations, these results provide strong empirical evidence that the stack representations are causally involved in the decoder's next token predictions.  The sharp collapse points to the absence of alternative encodings partially compensating for the ablation of the stack direction.

\begin{figure}[h]
\begin{center}
\includegraphics[width=0.8\linewidth]{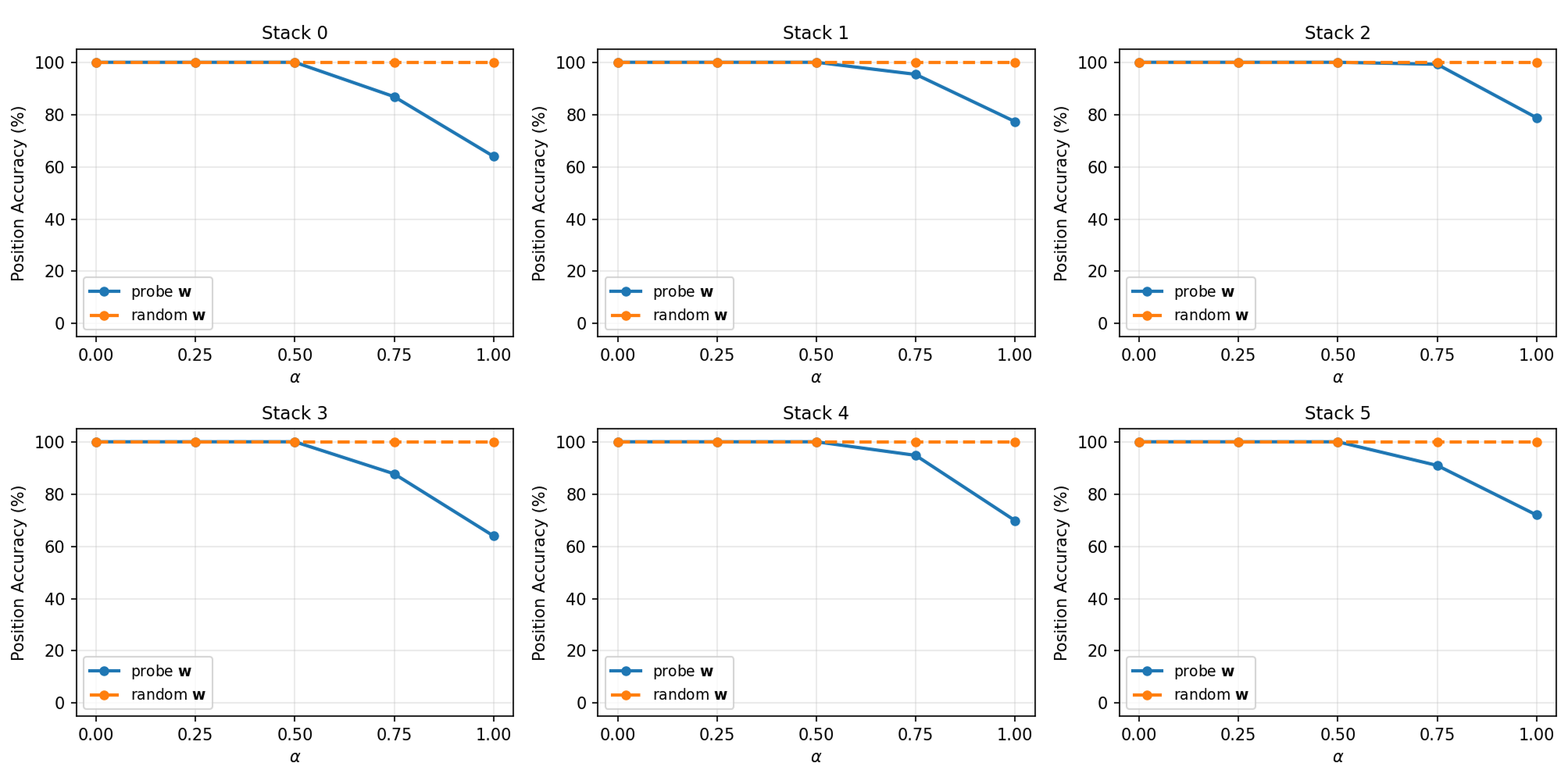}
\end{center}
\caption{Position accuracy as a function of ablation strength for $\alpha$ for Shuffle-6. Accuracy drops from targeted ablation, random ablation keeps accuracy intact.}
\label{fig:pos_acc}
\end{figure}

\section{Future Work}
\label{others}

A natural extension is to intervene \textit{mid-network}. Importantly, the ablation is applied only at inference time to the final hidden states, without re-running the transformer. This isolates the causal contribution of the representation to the decoder’s predictions. How the transformer tackles the manipulation of emergent computational primitives like stacks and counters while modeling formal languages needs to be explored by passing ablated or edited states through the feed-forward network and observing the change in outputs. Furthermore, the role of different architectures in the emergence of representations, their redundancy, causality, and their relationships remains an interesting direction.

\bibliography{iclr2026_conference}

@inproceedings{
tiwari2025emergent,
title={Emergent Stack Representations in Modeling Counter Languages Using Transformers},
author={Utkarsh Tiwari and Aviral Gupta and Michael Hahn},
booktitle={ICLR 2025 Workshop on World Models: Understanding, Modelling and Scaling},
year={2025},
url={https://openreview.net/forum?id=o5xTyDilNs}
}

@article{belinkov-2022-probing,
    title = "Probing Classifiers: Promises, Shortcomings, and Advances",
    author = "Belinkov, Yonatan",
    journal = "Computational Linguistics",
    volume = "48",
    number = "1",
    month = mar,
    year = "2022",
    address = "Cambridge, MA",
    publisher = "MIT Press",
    url = "https://aclanthology.org/2022.cl-1.7/",
    doi = "10.1162/coli_a_00422",
    pages = "207--219"
}

@misc{elhage2022toymodelssuperposition,
      title={Toy Models of Superposition}, 
      author={Nelson Elhage and Tristan Hume and Catherine Olsson and Nicholas Schiefer and Tom Henighan and Shauna Kravec and Zac Hatfield-Dodds and Robert Lasenby and Dawn Drain and Carol Chen and Roger Grosse and Sam McCandlish and Jared Kaplan and Dario Amodei and Martin Wattenberg and Christopher Olah},
      year={2022},
      eprint={2209.10652},
      archivePrefix={arXiv},
      primaryClass={cs.LG},
      url={https://arxiv.org/abs/2209.10652}, 
}

@inproceedings{hewitt-liang-2019-designing,
    title = "Designing and Interpreting Probes with Control Tasks",
    author = "Hewitt, John  and
      Liang, Percy",
    editor = "Inui, Kentaro  and
      Jiang, Jing  and
      Ng, Vincent  and
      Wan, Xiaojun",
    booktitle = "Proceedings of the 2019 Conference on Empirical Methods in Natural Language Processing and the 9th International Joint Conference on Natural Language Processing (EMNLP-IJCNLP)",
    month = nov,
    year = "2019",
    address = "Hong Kong, China",
    publisher = "Association for Computational Linguistics",
    url = "https://aclanthology.org/D19-1275/",
    doi = "10.18653/v1/D19-1275",
    pages = "2733--2743"
}

@article{hahn-2020-theoretical,
    title = "Theoretical Limitations of Self-Attention in Neural Sequence Models",
    author = "Hahn, Michael",
    editor = "Johnson, Mark  and
      Roark, Brian  and
      Nenkova, Ani",
    journal = "Transactions of the Association for Computational Linguistics",
    volume = "8",
    year = "2020",
    address = "Cambridge, MA",
    publisher = "MIT Press",
    url = "https://aclanthology.org/2020.tacl-1.11/",
    doi = "10.1162/tacl_a_00306",
    pages = "156--171"
}

@inproceedings{bhattamishra-etal-2020-ability,
    title = "On the {A}bility and {L}imitations of {T}ransformers to {R}ecognize {F}ormal {L}anguages",
    author = "Bhattamishra, Satwik  and
      Ahuja, Kabir  and
      Goyal, Navin",
    editor = "Webber, Bonnie  and
      Cohn, Trevor  and
      He, Yulan  and
      Liu, Yang",
    booktitle = "Proceedings of the 2020 Conference on Empirical Methods in Natural Language Processing (EMNLP)",
    month = nov,
    year = "2020",
    address = "Online",
    publisher = "Association for Computational Linguistics",
    url = "https://aclanthology.org/2020.emnlp-main.576/",
    doi = "10.18653/v1/2020.emnlp-main.576",
    pages = "7096--7116"
}

@article{DBLP:journals/corr/abs-1901-03429,
  author       = {Jorge P{\'{e}}rez and
                  Javier Marinkovic and
                  Pablo Barcel{\'{o}}},
  title        = {On the Turing Completeness of Modern Neural Network Architectures},
  journal      = {CoRR},
  volume       = {abs/1901.03429},
  year         = {2019},
  url          = {http://arxiv.org/abs/1901.03429},
  eprinttype    = {arXiv},
  eprint       = {1901.03429},
  bibsource    = {dblp computer science bibliography, https://dblp.org}
}

@misc{nanda2023progressmeasuresgrokkingmechanistic,
      title={Progress measures for grokking via mechanistic interpretability}, 
      author={Neel Nanda and Lawrence Chan and Tom Lieberum and Jess Smith and Jacob Steinhardt},
      year={2023},
      eprint={2301.05217},
      archivePrefix={arXiv},
      primaryClass={cs.LG},
      url={https://arxiv.org/abs/2301.05217}, 
}

@misc{meng2023locatingeditingfactualassociations,
      title={Locating and Editing Factual Associations in GPT}, 
      author={Kevin Meng and David Bau and Alex Andonian and Yonatan Belinkov},
      year={2023},
      eprint={2202.05262},
      archivePrefix={arXiv},
      primaryClass={cs.CL},
      url={https://arxiv.org/abs/2202.05262}, 
}
\bibliographystyle{iclr2026_conference}
\newpage
\appendix
\section{Appendix}
Experimental results for Shuffle-k for different values of \(k\).

\subsection{Shuffle-2}
\begin{figure}[h]
\begin{center}
\includegraphics[width=0.8\linewidth]{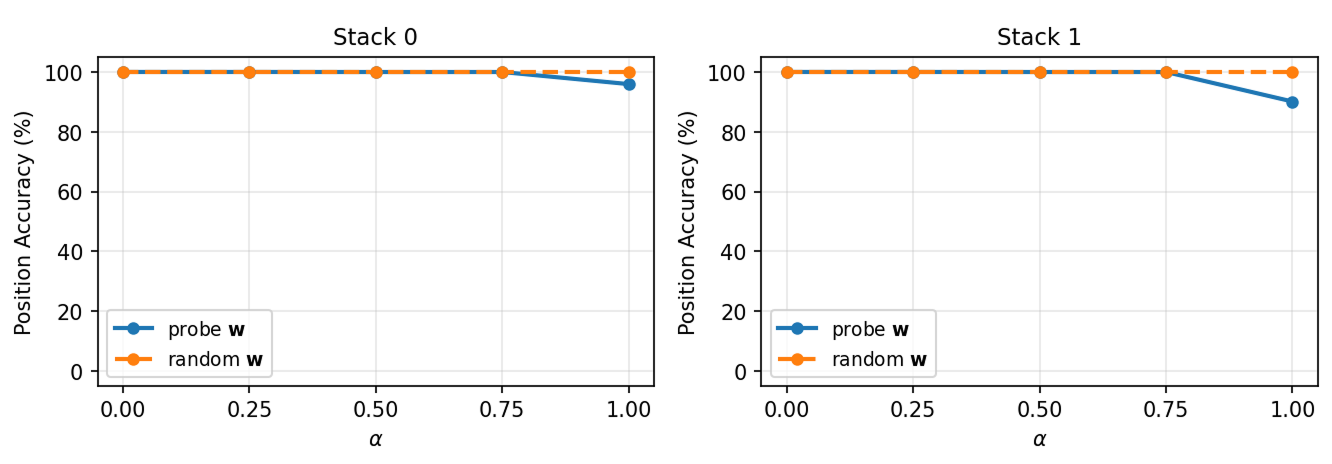}
\end{center}
\caption{Position accuracy as a function of ablation strength $\alpha$.}
\end{figure}

\begin{figure}[h]
\begin{center}
\includegraphics[width=0.8\linewidth]{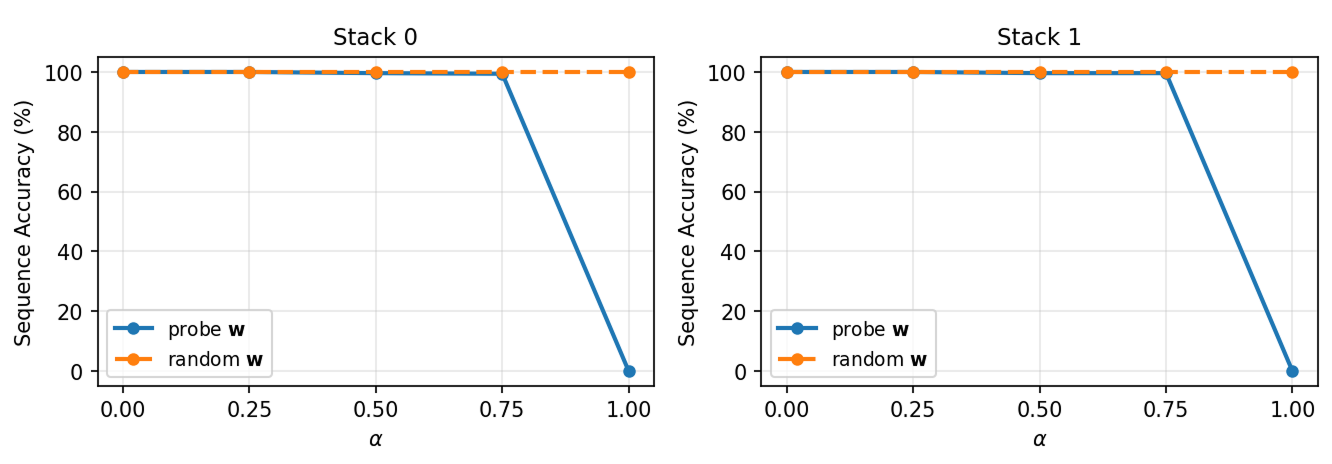}
\end{center}
\caption{Sequence accuracy as a function of ablation strength $\alpha$.}
\end{figure}
\newpage

\subsection{Shuffle-4}
\begin{figure}[h]
\begin{center}
\includegraphics[width=0.8\linewidth]{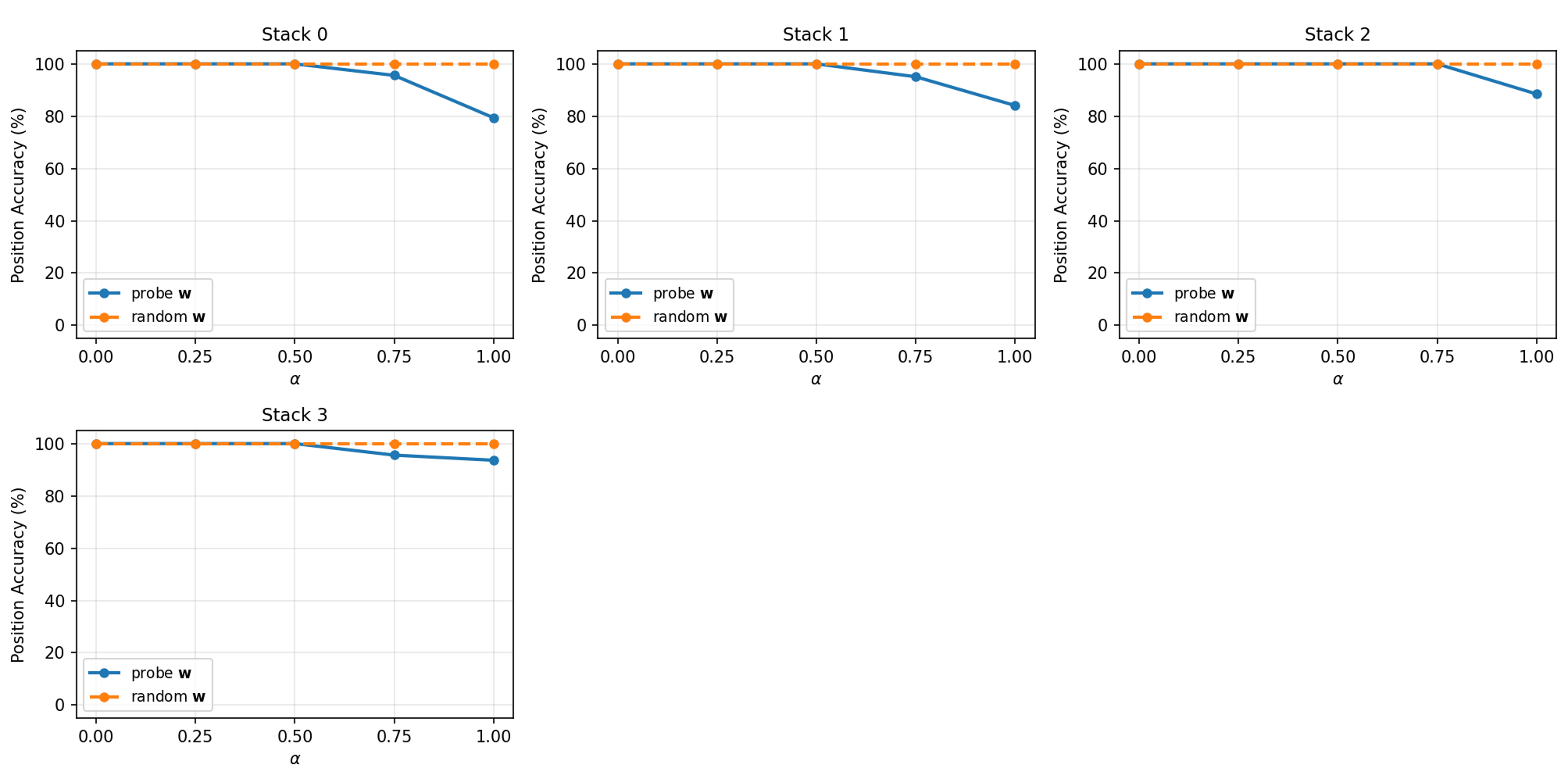}
\end{center}
\caption{Position accuracy as a function of ablation strength $\alpha$.}
\end{figure}

\begin{figure}[h]
\begin{center}
\includegraphics[width=0.8\linewidth]{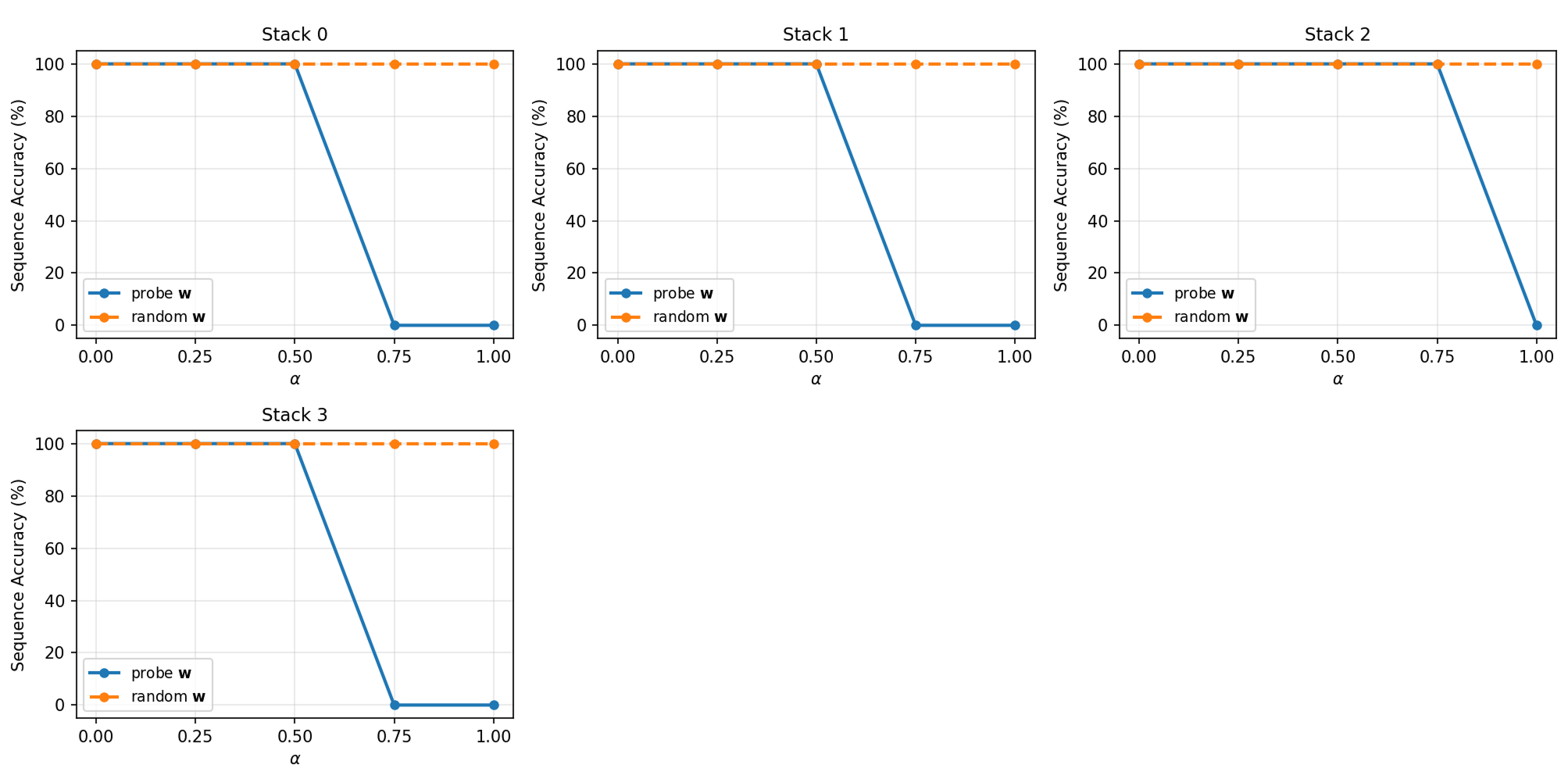}
\end{center}
\caption{Sequence accuracy as a function of ablation strength $\alpha$.}
\end{figure}
\newpage

\subsection{Shuffle-8}
\begin{figure}[h]
\begin{center}
\includegraphics[width=0.8\linewidth]{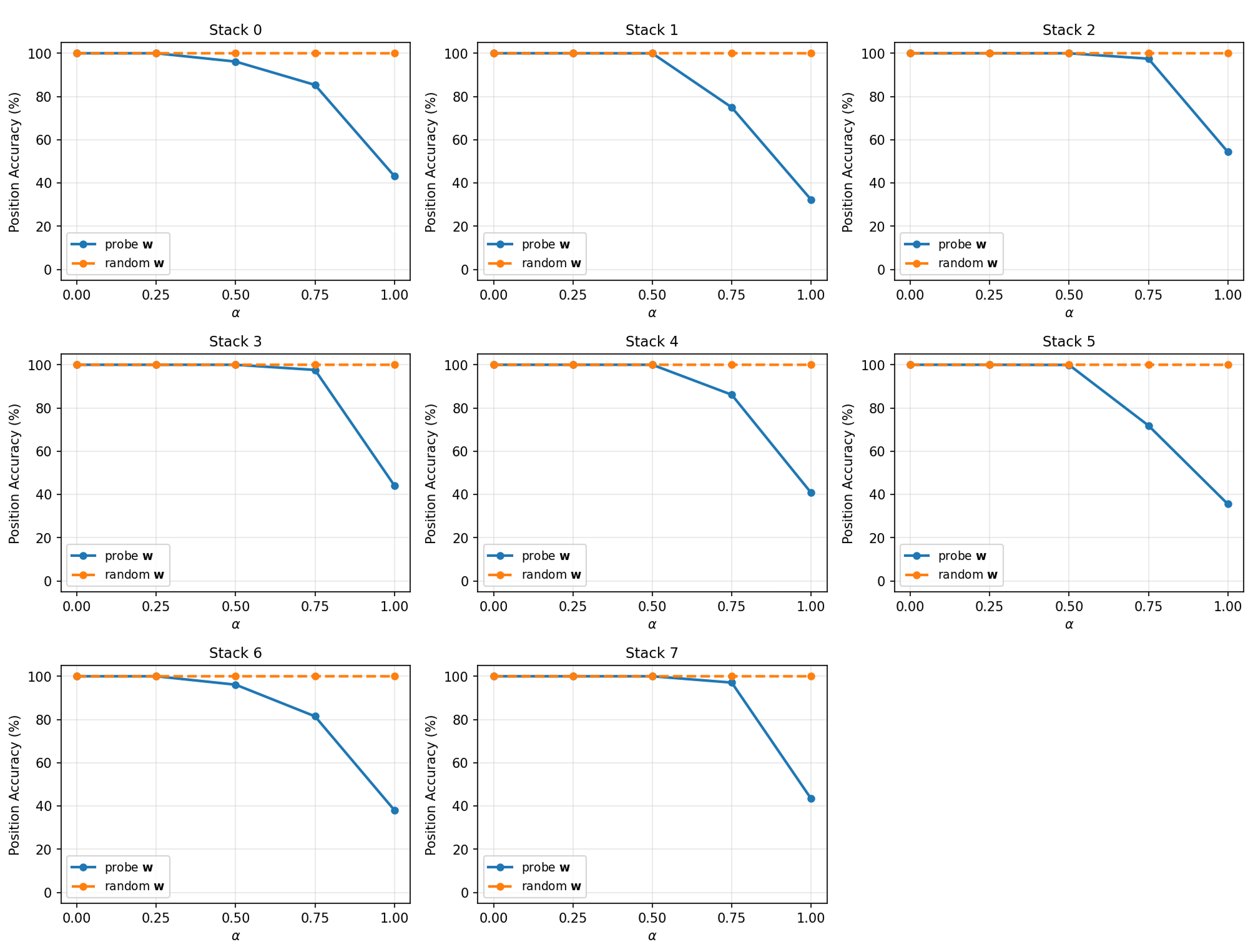}
\end{center}
\caption{Position accuracy as a function of ablation strength $\alpha$.}
\end{figure}

\begin{figure}[h]
\begin{center}
\includegraphics[width=0.8\linewidth]{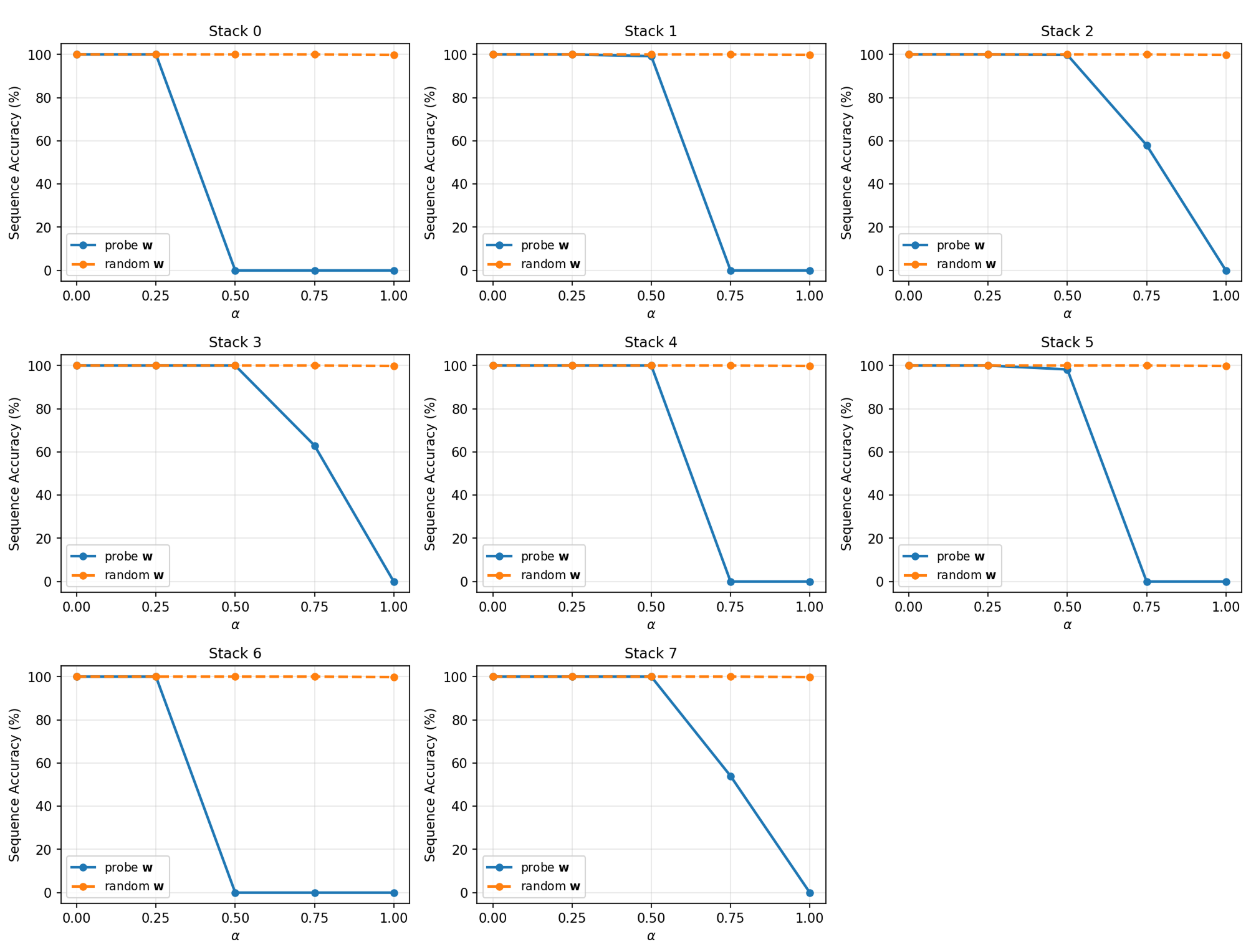}
\end{center}
\caption{Sequence accuracy as a function of ablation strength $\alpha$.}
\end{figure}

\end{document}